\documentclass[runningheads,a4paper]{llncs}

\usepackage{amssymb}
\usepackage{graphicx}
\usepackage{multirow}

\usepackage{url}
\urldef{\mailsa}\path|g.marra@unifi.it|
\urldef{\mailsb}\path|andrea.zugarini@unifi.it|
\urldef{\mailsc}\path|mela@diism.unisi.it| 
\urldef{\mailsd}\path|marco.maggini@unisi.it|
\newcommand{\keywords}[1]{\par\addvspace\baselineskip
\noindent\keywordname\enspace\ignorespaces#1}

\begin{document}

\mainmatter  

\title{An Unsupervised Character-Aware Neural Approach to Word and Context Representation Learning\thanks{This is a post-peer-review, pre-copyedit version of an article published in LNCS, volume 11141. The final authenticated version is available online at: \texttt{https://doi.org/10.1007/978-3-030-01424-7\_13}}}
\titlerunning{A Character-Aware Model to Word and Context Representation Learning}

\author{Giuseppe Marra \inst{1 \and 2}\and Andrea Zugarini \inst{1 \and 2}\and Stefano Melacci \inst{2}\and Marco Maggini \inst{2}}
\authorrunning{G. Marra et al.}

\institute{
	DINFO, University of Firenze, Italy\\ 
\and
	DIISM, University of Siena, Italy\\
	\mailsa, \mailsb, \mailsc, \mailsd\\
}

%
%

\maketitle

\begin{abstract}
In the last few years, neural networks have been intensively used to develop meaningful distributed representations of words and contexts around them.
When these representations, also known as ``embeddings'', are learned from unsupervised large corpora, they can be transferred to different tasks with positive effects in terms of performances, especially when only a few supervisions are available. In this work, we further extend this concept, and we present an unsupervised neural architecture that jointly learns word and context embeddings, processing words as sequences of characters. This allows our model to spot the regularities that are due to the word morphology, and to avoid the need of a fixed-sized input vocabulary of words. We show that we can learn compact encoders that, despite the relatively small number of parameters, reach high-level performances in downstream tasks, comparing them with related state-of-the-art approaches or with fully supervised methods.

\end{abstract}

\section{Introduction}
\label{sec:intro}
Recent advances in Natural Language Processing (NLP) are characterized by the development of techniques that compute powerful word embeddings and 
by the extensive use of neural language models. Word Embeddings (WEs) aim at representing individual words in a
low--dimensional continuous space, in order to exploit its topological properties to model semantic or grammatical
relationships between different words. In particular, they are based on the assumption that functionally or semantically
related words appear in similar contexts. 

Despite the idea of continuous word representations was proposed a several years ago \cite{hinton1986distributed}, their importance became strongly popular mostly after the work of Mikolov et al. \cite{word2vec}, when the CBOW and Skip--Gram models were introduced as implementations of the {\em word2vec} idea.
Key features of these models are the unsupervised scheme of the learning process and the simplicity of the computation that allows a highly efficient training from very large unlabeled corpora. Moreover, the learning objective function is task--independent, such that it allows the development of embeddings suitable for several NLP tasks.
WEs are generally constituted by a single vector to represent each specific word in a vocabulary $V$ of $N=|V|$ words.
The requirement of a predefined vocabulary is an important limitation for every NLP model. Rare and Out--Of--Vocabulary (OOV) words
will not have a meaningful vector representation. Moreover, WEs do not take into account morphological properties of words. For instance,
the same suffix \emph{ing} may suggest that two words have some functional similarity. Hence, the  information conveyed by the sequence of characters
representing a word may be useful to tackle both the problem of unseen words and the modelling of morphology for in--vocabulary tokens.
For instance, the character structure of tokens can also help to detect Named Entities, usually treated as OOV elements, recognizing proper nouns, by means of capital letters, or acronyms. Furthermore, a character--based model can deal with noise caused by typos, slang, etc, that are common issues in open--domain systems such as conversational agents or sentiment analysis tools. 

There are several NLP tasks in which it is useful to generate vectorial representations of contexts too. In fact, polysemy and homonymy cause inherent semantic ambiguities in language interpretation, that can only be resolved by looking at the surrounding context, that is the goal of the Word Sense Disambiguation (WSD) task.
Neural approaches have been developed to learn context embeddings, such as \emph{context2vec} \cite{context2vec}.

In this work we propose a character--based unsupervised model to learn both context and word embeddings from generic text. The model consists in a hierarchy of two distinct Bidirectional Long Short Term Memories (Bi--LSTMs) \cite{bidirectional}, to encode words as sequences of characters and word--level contextual representations, respectively. 
Our unsupervised learning approach, despite being more compact than other related algorithms, yields generic embeddings with features that 
can be efficiently exploited in different NLP tasks requiring either word or context embeddings, such as chunking and WSD, as we show in our comparisons.

The paper is structured as follows. First, in Section \ref{sec:related_work} the related work is summarized. Then, we describe the proposed model in Section \ref{sec:model}. Section \ref{sec:experiments} reports our experimental results and Section \ref{sec:conclusions} draws our conclusions and the directions for future work.

\section{Related Work}
\label{sec:related_work}
Our unsupervised computational scheme follows the one of the CBOW instance of the \emph{word2vec} algorithm \cite{word2vec}. 
The method we propose in this paper is inspired by the ideas behind \emph{context2vec} \cite{context2vec}, that we extend with a bidirectional recurrent neural model that processes words as sequence of characters. We also focus on a single encoder that we use both to represent words alone and words belonging to a context.

There are several approaches that jointly learn task-oriented (supervised) word and character--based representations, that are subsequently either concatenated or combined by a non--linear function. In \cite{gatedwordchar} a gate adaptively decides how to mix
the two representations, whereas the models proposed in \cite{charPOS} and \cite{charNER} exploit the concatenation of word embeddings and character representations to address  Part--Of--Speech (POS) Tagging and Named Entity Recognition (NER), respectively. Differently, our work focusses on a single character-level encoder that is trained in an unsupervised manner.

There exists a number of different approaches that extract vectorial representations directly from the character sequences of words, mostly focused on Language Modeling (LM) or Character Language Modeling (CLM).  These representations are generally computed by either Convolutional Neural Networks (CNNs) or Recurrent Neural Networks (RNNs) - mostly LSTMs \cite{lstms}. Ling et al. \cite{ling2015finding} applied Bidirectional LSTMs \cite{bidirectional} to learn task--dependent character level features for Language Modeling and POS tagging, showing particular improvements in morphologically rich languages such as Turkish. A multi--layered Hierarchical Recurrent Neural Network was applied in \cite{hierachicalcharlm} to solve CLM. Differently from our approach, the output of this model is a distribution over characters, while we exploit word level predictions. The character--aware model of \cite{charaware}, is based on a highway--network on top of 1-d convolutional filters processing input characters. The resulting output is then handled by a LSTM for a LM task. The highway--network output does provide the distributed representation of a word.  
In \cite{exploringlimitslm} different architectures, mostly based on CNNs, are studied in LM tasks.
The proposed approach differs from most of the previous ones (1) for the learning mechanism, that is completely unsupervised on large text corpora, thus allowing the development of task--independent representations, and (2) for the architecture that is aimed at obtaining character--aware representations of both contexts and words, that are suitable for a large variety of NLP applications.

\section{The Character--Aware Neural Model}
\label{sec:model}

The proposed model is organized as a hierarchical architecture based on Bi--LSTMs processing sentences. Each sentence is first split into a sequence of words using space characters (i.e. whitespaces, tabs, newlines, etc. ) as separators. Words are further split into sequences of characters, such
that there is no need to specify a vocabulary in advance. Then, the character sequence of an input word $x$ is processed to obtain its vectorial representation (word embedding), while the character sequences of the surrounding words are used to encode the context to which $x$ belongs (context embedding). Given the current sentence, the context of $x$ comprises the words that precede and follow $x$. Inspired by the CBOW scheme \cite{word2vec}, our model is trained to predict the current word given its context. In the following we describe each layer of the proposed architecture.

\subsection{Word and context embeddings}
We consider an input sentence $s$ composed of $n$ words, $s = (x_1, \ldots, x_n)$, where each word is a sequence of characters $x_i = (c_{i,1}, \ldots ,c_{i,|x_i|})$, being $|x_i|$ the length of the sequence $x_i$. Each character $c_{ij}$ is encoded as an index in a dictionary of $C$ characters and it is mapped to a real vector $\hat{c}_{ij} \in \mathbb{R}^{d_c}$ as
\begin{equation}
\label{eq:char_emb}
\hat{c}_{ij} = W_c \cdot 1(c_{ij}),
\end{equation}
where $W_c \in \mathbb{R}^{C \times d_c}$ is the matrix of the learnable character representations, each of them of size $d_c$, while $1(\cdot)$ is a function returning a one-hot representation of its integer input. Note that $C$ is quite small, in the order of hundreds, compared to common word vocabularies, whose size is in the order of hundreds of thousands.

For each input word $x_i$,  the first layer of the model extracts a \textit{word embedding} $e_i $, using a bidirectional recurrent neural network with LSTM cells (Bi-LSTM) \cite{biDirGraves}.  Let $\overrightarrow{{r}_{c}}$ and $\overleftarrow{{r}_{c}}$ be the forward and backward components of a Bi-LSTM taking a sequence of character embeddings as input and returning their internal states $\overrightarrow{h_c}$ and $\overleftarrow{h_c}$ after the entire sequence has been processed. The embeddings $e_i $ of the word $x_i$ is then the concatenation of $\overrightarrow{h_c}$ and $\overleftarrow{h_c}$:
\begin{equation}
\label{unos}
e_i =[\overrightarrow{h_c}, \overleftarrow{h_c}] =[ \overrightarrow{{r}_{\hat{c}}}(\hat{c}_{i,1} \ldots \hat{c}_{i,|x_i|}), \overleftarrow{{r}_{c}}(\hat{c}_{i,|x_i|} \ldots \hat{c}_{i,1})], 
\label{eq:WE}\\
\end{equation}
where we indicated with $[ \cdot \:, \cdot ]$ the concatenation operation and we emphasized the backward nature of $\overleftarrow{{r}_{c}}$ by showing the character sequence in reverse order.  

The second layer follows a similar scheme to compute the \textit{contextual embedding} $\hat{e}_i$ of the word $x_i$ in the sentence $s$.  Let $\overrightarrow{{r}_{e}}$ and $\overleftarrow{{r}_{e}}$ be the forward and backward components of a Bi-LSTM taking as inputs the embeddings of left context of $x_i$ (i.e $[e_1, \ldots, e_{i-1}]$) and of the right context of $x_i$ (i.e $[e_{i+1}, \ldots, e_n]$), respectively. Given the Bi-LSTM internal states $\overrightarrow{h_e}$ and $\overleftarrow{h_e}$ obtained after processing the input left and right context sequences, the contextual embedding $\hat{e_i} $ of the word $x_i$ is then obtained by projecting the concatenation of $\overrightarrow{h_e}$ and $\overleftarrow{h_e}$ into a lower-dimensional space by means of a Multi-Layer Perceptron (MLP), with the goal of merging and compressing the left and right context representations,
\begin{equation}
\label{dos}
\hat{e}_i = MLP([\overrightarrow{h_e}, \overleftarrow{h_e}])= MLP([\overrightarrow{{r}_{e}}(e_{1} \ldots, e_{i - 1}), \overleftarrow{{r}_{e}}(e_{n}\ldots e_{i+1})]).
\label{eq:hatCWE}
\end{equation}

\begin{figure}[!h]
	\label{fig:architecture}
	\centering
	\includegraphics[scale=0.31]{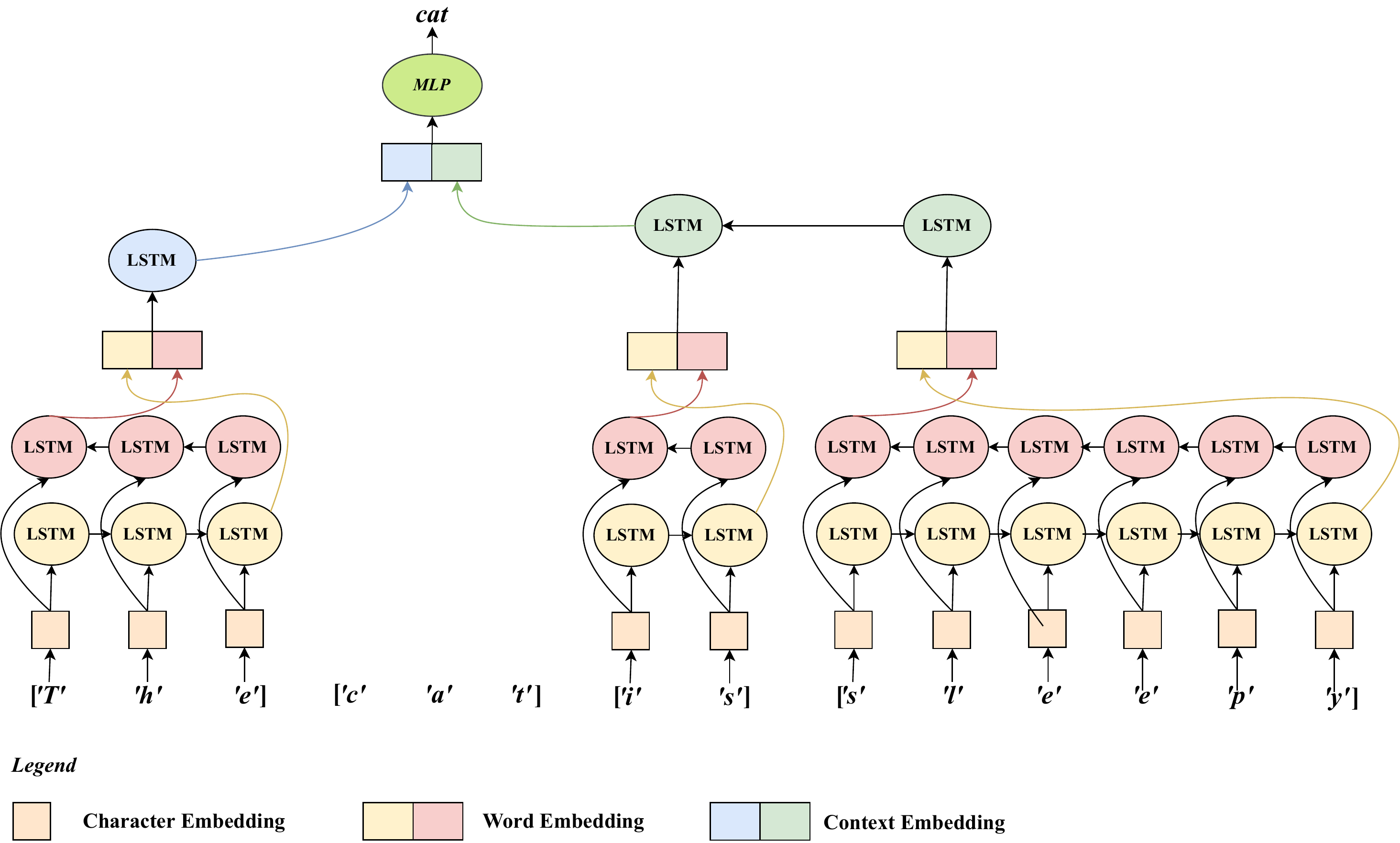}
	\caption[Word and Contextual Embeddings]{The sentence ``\textit{The cat is sleepy}'' is fed to our model, with target word \textit{cat}. The sequence of character embeddings (orange squares on the bottom) are processed by the word--level Bi-LSTM yielding the word embeddings (yellow--red squares in the middle). The context-level Bi-LSTM processes the word embeddings in the left and right contexts of \textit{cat}, to compute a representation of the whole context (blue--green squares on the top). Such representation is used to predict the target word \textit{cat}, after having projected it by means of a MLP.}
\end{figure}

The overall architecture is sketched in Figure~\ref{fig:architecture}.
Notice that $e_i$ is the embedding of word $x_i$, whereas $\hat{e}_i$ is the representation of $x_i$ in the context of $s$ without including $x_i$ itself. Hence, the model computes at the same time word (Eq. (\ref{eq:WE})) and context (Eq. (\ref{eq:hatCWE})) embeddings for a specific word. 

\subsection{Learning algorithm}\label{sec:learning}
Both word and context representations are learned following the unsupervised approach used in CBOW \cite{context2vec,word2vec}. Given a corpus of textual data, the objective of our model is to predict each word given the representation of its surrounding context (Eq. (\ref{eq:hatCWE})). 
In particular, the context embedding of Eq. (\ref{eq:hatCWE}) is projected into the space of the corpus vocabulary using a linear projection. Instead of performing a softmax activation and minimizing the cross-entropy (as commonly done in LM tasks), the whole network is trained by minimizing the Noise Contrastive Estimation (NCE) loss function \cite{nce}.  
%
NCE belongs to a family of classification algorithms, which approximate a softmax regression by means of sampling methods. NCE is particularly helpful in all those cases in which the number of output units is prohibitively high, as it is for our (and related) model. 

One could argue that a vocabulary of words is still needed, since it is required to make the aforementioned word prediction. However, this is not a limitation, since it is only necessary at training time, while it is not needed when deploying the model. In principle, a different approach would be feasible, where the context representation of Eq. (\ref{eq:hatCWE})  is decoded into a sequence of characters that represent the word to predict. We tried both approaches and we found the word level prediction to give the best results. Thanks to the dynamic behaviour of the context-level RNNs, our model can deal with contexts of any length. In this work, the state of the RNN $r_e$ is reset at the beginning of a new sentence, to reduce the variability of the contexts.


\section{Experimental Results}
\label{sec:experiments}
\begin{figure}[!h]
	\centerline{
	\includegraphics[scale=0.33]{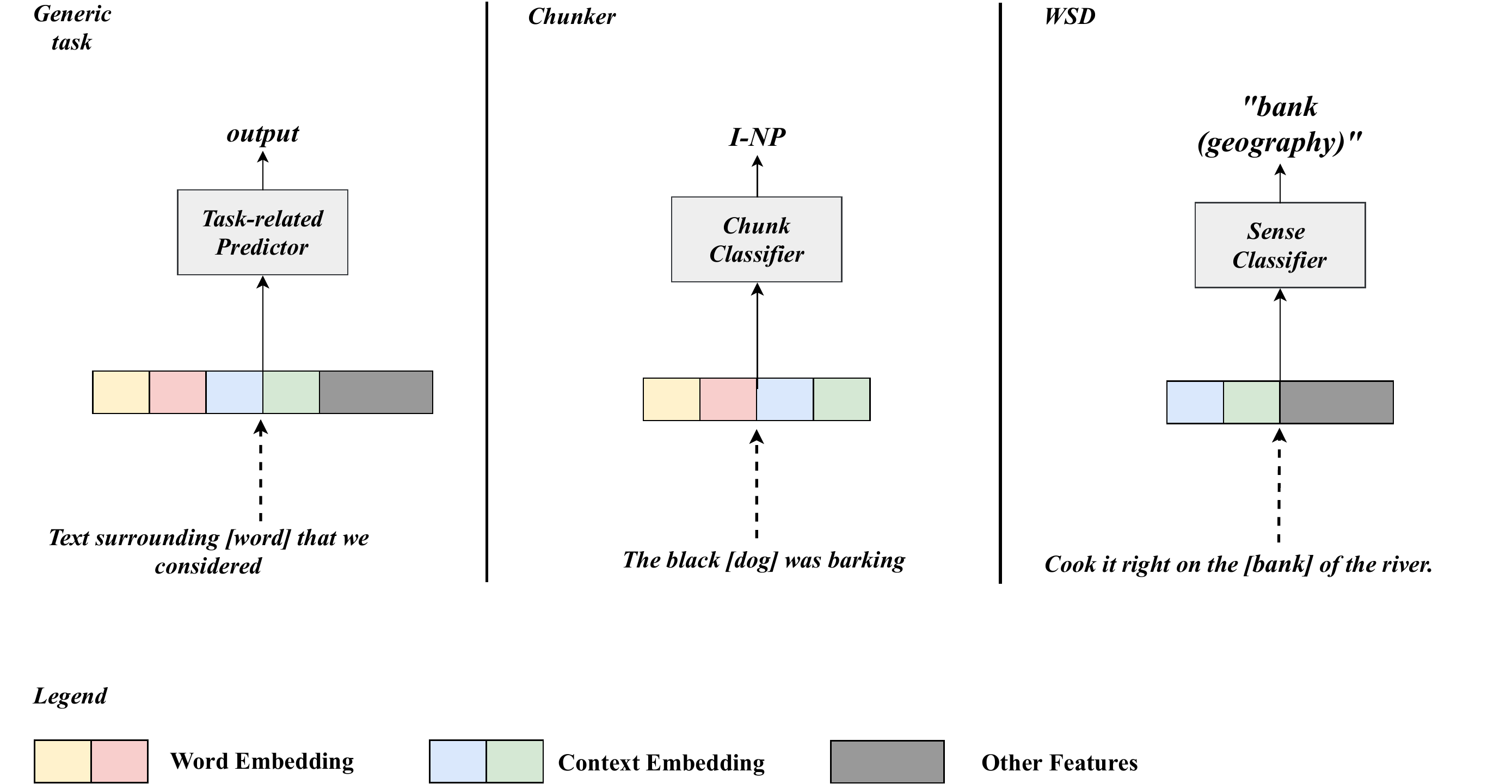}}
	\caption{Examples of how word and context embeddings can be used in a generic task, and in the cases of Chunking and WSD of this paper.}
	\label{fig:classifiers}
\end{figure}

We conducted different experiments to evaluate the word and context representations developed by the proposed model. In particular, we first trained our model on a large corpora. Then, we detached the learned word and context encoders and considered the tasks of Chunking and Word Sense Disambiguation (WSD), exploiting our word and context embeddings as features for each task-specific classifier, as shown in Figure \ref{fig:classifiers}. Depending on the problem at hand, it may be useful to use either both the word and context embeddings or only one of them. Any other additional features can also be concatenated to these representations to obtain a richer input vector.
We also evaluated the robustness of our model to character--level noise. Hence, we considered the WSD task when the input words are perturbed by typos modelled as random replacements of single characters.
Finally we report some qualitative examples, showing the nearest neighbours for both word and context representations of a set of sample words.

\textbf{Model setup}.
Our model has been trained on the ukWaC corpus\footnote[1]{\url{http://wacky.sslmit.unibo.it/doku.php?id=corpora}} (2 billion words). The size $d_c$ of the character embeddings is set to 50, whereas
word and context embeddings are of sizes 1000 and 600, respectively. The MLP, that maps the RNN states into the context embeddings, has one
hidden layer of 1200 units with ReLU activation functions. These settings are inspired by those used in the \emph{context2vec} architecture \cite{context2vec} (the structure of the last projection layer described in Subsection \ref{sec:learning} is the same).
The complete encoding model has around 7 million trainable parameters, which is about \textsc{16 times smaller} than the \textit{context2vec} model in \cite{context2vec}; this is due to the fact that words are encoded using a RNN that does not depend on the vocabulary size.

\textbf{Chunking}.
Chunking is a classical NLP problem whose goal is to tag text segments with labels defining their syntactic roles, e.g. noun phrase (NP) or verbal phrase (VP).
Each word is uniquely associated with a single tag expressing the segment class and its position within the phrase. An instance of Chunking classification is shown in Figure \ref{fig:classifiers}, where the word \emph{dog} is marked with the label \emph{I-NP}, standing for Inside-chunk Noun Phrase.
A standard benchmark for Chunking is the CoNLL 2000 dataset that contains 211,727 tokens in the training set and 47,377 tokens in the test set. 
The chunk tag is predicted by training a classifier that receives as input only the concatenation of the word and context embeddings computed by the model.
This vector is projected onto a 600 dimensional space, and further processed by a Bi-LSTM that outputs vectors of size 500 that are finally mapped to the space
of 23 classes, representing the chunk tags.
Weights are updated using Adam Optimizer with default hyper-parameters and weight decay regularization with a factor of $0.001$. 
\begin{table}[h]
	\caption{Results on the Chunking task - different input features.}
	\centerline{
	\begin{tabular}{lc}
		\hline
		Input Features		& F1 $\%$ \\ \hline
	  Our WE only	& 89.68  \\
	Our CE only	& 89.59 \\
	Our WE + Our CE	&  \textbf{93.30}\\ 
	WE + CE Trained on Task&  89.83\\\hline
	\end{tabular}}
	\label{tab:chunking_exp}
\end{table}
We compared several variants of the proposed model and the resulting F1 scores are shown in Table \ref{tab:chunking_exp}. We report results when using only Word Embeddings (WE),
only Context Embeddings (CE), and both of them (WE+CE). In this case we also considered WE and CE that are not generated by our model, but that are variables of the whole architecture trained with the task-level supervision. Both the feature types (WE and CE) are needed to achieve better performances, as expected. This experiment highlights the importance of using embeddings that are pre-trained with our model, that allows us to obtain the best F1 score of $93.30$. This value can be compared with the results reported by Collobert et al. \cite{nlpfromscratch} (94.32) and  
by Huang et al. \cite{huang2015} (94.46), taking into account that in our case we did not make use of any hand-crafted feature nor of any kind of post-processing to adjust incoherent predictions.
Moreover, when adding POS tagging features, our model reaches the same performances (93.94) of the state-of-the-art architecture \cite{huang2015} without Conditional Random Fields.
Hence, we can conclude that the proposed architecture provides word and context embeddings that convey enough information to reach competitive performances.  Furthermore, it should be considered that the number of parameters in the model is dramatically reduced with respect to such competitors, since there is no word vocabulary.

\textbf{Word Sense Disambiguation}.
Experiments on WSD were carried out within the evaluation framework proposed in \cite{raganato2017}, that collects multiple benchmarks (Senseval*, SemEval*, and a merged collection - ALL). The goal of WSD is to identify the correct sense of words. 
We followed the commonly used IMS approach \cite{ims}, that is based on an SVM classifier on top of conventional WSD features. We compare our method against the original IMS model and other instances of it in which the WSD features are augmented with different context embeddings. 
\begin{table}[!h]
	\caption{Word Sense Disambiguation in the benchmarks collected in \cite{raganato2017}. The best results (F1 \%) are obtained by the {\em contex2Vec} model that however has \textsc{16 times more parameters
	  than the proposed model} and no capability to deal with OOV tokens.}
	  \centerline{
	\begin{tabular}{l|ccccccc}
		\hline
		Model & Senseval2 & Senseval3 & SemEval2007 & SemEval2013 & SemEval2015& ALL\\ \hline
		IMS & 70.2 & 68.8 & 62.2 & 65.3 & 69.3 & 68.1  \\
		IMS+word2vec &  72.2& 69.9& 62.9& 66.2& 71.9& 69.6  \\
		IMS+context2vec & \textbf{73.8} & \textbf{71.9} & \textbf{63.3} & \textbf{68.1} & \textbf{72.7} & \textbf{71.1}  \\
		IMS+Our CE & 72.8 & 70.5 & 62.0 & 66.2 & 71.9 & 69.9 \\ \hline
	\end{tabular}}
	\label{tab:wsd_exp}
\end{table}
We report the results in Table \ref{tab:wsd_exp} and \ref{tab:wsd_pos_exp}. Our embeddings outperform both the IMS with only conventional features and {\em word2vec} embeddings, opportunely averaged \cite{webis}, moreover it is competitive with \emph{context2vec} representations. It is also worth to mention that, to the best of our knowledge, the use of \emph{context2vec} features as input of the IMS is a novel attempt in the literature.
\begin{table}[!h]
	\caption{Overall results (F1 \%) grouped by Part of Speech (ALL benchmark \cite{raganato2017}).}
	\centerline{
	\begin{tabular}{l|cccc}
		\hline
		Model & Noun & Adjective & Verb & Adverb  \\ \hline
		IMS & 70.0 & 75.2 & 56.0 & 83.2  \\
		IMS+word2vec	& 71.8 & 76.1 & 57.4 & 83.5  \\
		IMS+context2vec	&  \textbf{73.1} & \textbf{77.0} & \textbf{60.5} & 83.5  \\
		IMS+Our CE	&  71.3 & 76.6 & 58.1 & \textbf{83.8} \\ \hline
	\end{tabular}}
	\label{tab:wsd_pos_exp}
\end{table}

\textbf{Robustness to typos}.
Many NLP applications should deal with noisy textual data. Indeed, misspelled words are likely to be set as OOV in models based on word dictionaries. 
We compare the proposed model against \emph{context2vec} on a WSD task (ALL benchmark), when introducing an increasing probability to randomly perturb a character of a word.
\begin{figure}[!h]
	\centerline{
	\includegraphics[scale=0.26]{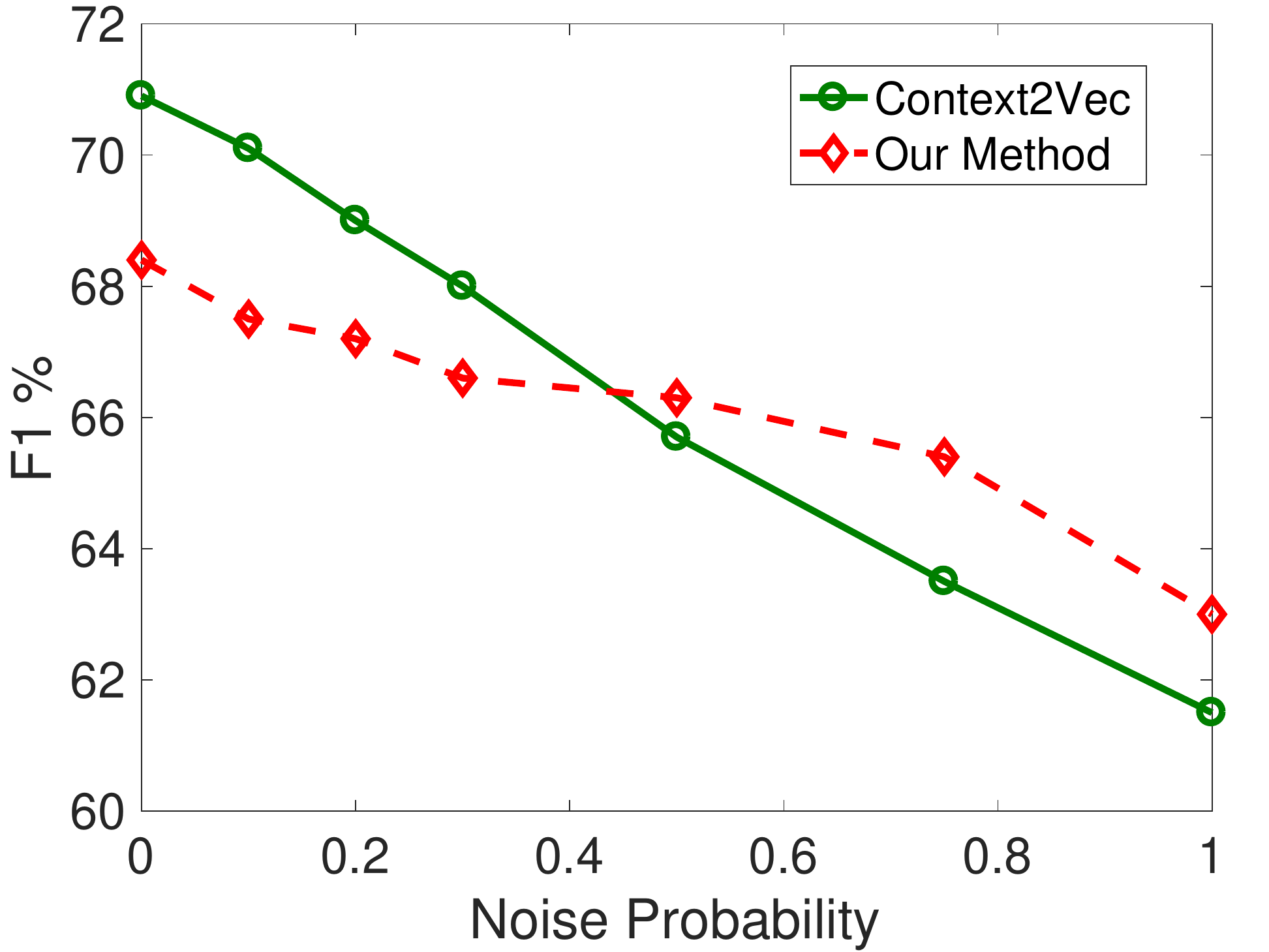}}
	\caption{Robustness to typos in a WSD task (ALL benchmark \cite{raganato2017}). The ``noise probability'' represents the probability of having a typo in a word.}
	\label{fig:noise_exp}
	\end{figure}
Conventional WSD features are completely removed for both the models, that only use context-level representations. Figure \ref{fig:noise_exp} shows how the F1 score decreases with the increase of the noise probability. Both the models suffer for word perturbations, but the character-aware embeddings yield a slower degradation in performances, that allows it to outperforms \emph{context2vec} for high levels of noise. 

\textbf{Qualitative evaluation}.
One of the most intriguing properties of embeddings is their capability to capture semantic and syntactic similarities into the topology of the embedding space. 
Such characteristic is illustrated by means of examples for both the representations (word and context) obtained by the proposed model. Distance between the distributed representations are computed by the cosine similarity.
In Table \ref{tab:we_knn} we show the 5 nearest neighbours for some given words. The examples show that the character based model is capable of capturing both morphological and semantic similarities.
\begin{table}[!h]
	\caption{Top-5 closest words for a given target word.}
	\centerline{
	\begin{tabular}{lllll}
		\hline
		& \emph{turkish} & \emph{sometimes}  & \emph{usually} & \emph{happiness}  \\ \hline
		& danish & somehow  & normally & weirdness \\
		& welsh & altogether & basically & fairness \\
		& french & perhaps & barely & deformity \\
		& kurdish & nonetheless & typically &  ripeness\\
		& swedish & heretofore & formerly & smoothness\\
		\hline
	\end{tabular}}
	\label{tab:we_knn}
\end{table}

For the evaluation of context representations, we considered 8 sentences related to 2 different topics (4 sentences each): capitals of states and pizza. A context embedding is obtained by considering the tokens around the word \emph{capital} or \emph{pizza}. Then, a random sentence is chosen as query, and the remaining sentences are sorted according to the distance between the query context embedding and their vectors. An example is shown in Table \ref{tab:ce_knn}, where it is clear that all the contexts related to {\em pizza} instances are closer to the query than sentences concerning {\em capitals}.
\begin{table}[!h]
	\caption{Some contexts sorted by descending cosine similarity with respect to the query context \textit{``I like eating [ ] with cheese and ham''} of (unused) target word \textit{pizza}.}
	\centerline{
	\begin{tabular}{p{3cm}|l|l}
		\hline
		 & Query: \textit{I like eating [ ] with cheese and ham.}                   & pizza  \\ \hline
		\multirow{7}{*}{\parbox{25mm}{Contexts sorted by descending cosine similarity}} & Do you like to eat [ ] with cheese and salami ? & pizza \\ 
		& Did you eat [ ] at lunch ? & pizza \\
		& What is the best [ ] i can eat here ? &  pizza\\
		& Paris is the [ ] and most populous city in France ... . &  capital\\
		& London is the  [ ] and most populous city of England ... . &  capital\\
		& Rome is the [ ] of Italian Republic . &  capital\\
		& Washington , D.C. , .... , is the [ ] of the United States . & capital\\ \hline
	\end{tabular}}
	\label{tab:ce_knn}
\end{table}


\section{Conclusions}
\label{sec:conclusions}
We presented an unsupervised neural model that can develop task-independent word and context representations using character-level inputs. We trained our model on a 2 billion word corpus, and the resulting word and context encoders were used to produce robust input features to approach some popular NLP tasks (Chunking, WSD). The proposed model has shown the capability of building powerful representations that are competitive to state-of-the-art embeddings generated by models with a significantly larger number of parameters. Our future work will include applications of this model to conversational systems.

\bibliographystyle{splncs03}
\bibliography{ICANN2018}

\end{document}